# QuaterNet: A Quaternion-based Recurrent Model for Human Motion

Dario Pavllo[1, 2]*
dario.pavllo@epfl.ch

David Grangier[2]
http://david.grangier.info

Michael Auli[2]

[1] École Polytechnique
Fédérale de Lausanne (EPFL)
Lausanne, Switzerland

[2] Facebook AI Research (FAIR)
Menlo Park, USA

**Abstract**

Deep learning for predicting or generating 3D human pose sequences is an active research area. Previous work regresses either joint rotations or joint positions. The former strategy is prone to error accumulation along the kinematic chain, as well as discontinuities when using Euler angle or exponential map parameterizations. The latter requires re-projection onto skeleton constraints to avoid bone stretching and invalid configurations. This work addresses both limitations. Our recurrent network, *QuaterNet*, represents rotations with quaternions and our loss function performs forward kinematics on a skeleton to penalize absolute position errors instead of angle errors. On short-term predictions, *QuaterNet* improves the state-of-the-art quantitatively. For long-term generation, our approach is qualitatively judged as realistic as recent neural strategies from the graphics literature.

## 1 Introduction

Modeling human motion is essential for many applications, including action recognition [12, 34], action detection [49] and computer graphics [22]. The prediction of sequences of joint positions of a 3D-skeleton has recently been addressed with neural networks, both for short-term [14, 37] and long-term predictions [22, 23]. Neural approaches have been very successful in other pattern recognition tasks [5, 20, 29]. Human motion is a stochastic sequential process with a high-level of intrinsic uncertainty. Given an observed sequence of poses, a rich set of future pose sequences are likely. Therefore, even with an excellent model, the intrinsic uncertainty implies that, when predicting a long sequence of future poses, predictions far in the future are unlikely to match a reference recording. Consequently, the literature often distinguish short and long-term prediction tasks. Short-term tasks are often referred to as *prediction* tasks and can be assessed quantitatively by comparing the prediction to a reference recording through a distance metric. Long-term tasks are often referred to as *generation* tasks and are harder to assess quantitatively. In that case, human evaluation is crucial.

This work addresses both short and long-term tasks, with the goal to match or exceed the state-of-the-art methods of the computer vision literature for short-term prediction and to match or exceed the state-of-the-art methods of the computer graphics literature for long-term generation. With that objective, we identify the limitations of current strategies and address them. Our contribution is

---

twofold. First, we propose a methodology for employing quaternions with recurrent neural networks. Other parameterizations, such as Euler angles, suffer from discontinuities and singularities, which can lead to exploding gradients and difficulty in training the model. Previous work tried to mitigate these issues by switching to *exponential maps* (also referred to as *axis-angle representation*), which makes them less likely to exhibit these issues but does not solve them entirely [17]. Second, we propose a differentiable loss function which conducts forward kinematics on a parameterized skeleton, and combines the advantages of joint orientation prediction with those of a position-based loss.

Our experimental results improve the state-of-the-art on angle prediction errors for short-term prediction on the Human3.6m benchmark. We also compare long-term generation quality with recent work from the computer graphics literature through human judgment. On this task, we match the quality of previous work on locomotion, while allowing on-line generation, and better control over the timings and trajectory constraints imposed by the artist.

The remainder of the paper examines related work (Section 2), describes our *QuaterNet* method (Section 3) and presents our experiments (Section 4). Finally, we draw some conclusions and delineate potential future work (Section 5). We also release our code and pre-trained models publicly at <https://github.com/facebookresearch/QuaterNet> .

## 2   Related Work

The modeling of human motion relies on data from motion capture. This technology acquires sequences of 3-dimensional joint positions at high frame rate (120 Hz – 1 kHz) and enables a wide range of applications, such as performance animation in movies and video games, and motion generation. In that context, the task of generating human motion sequences has been addressed with different strategies ranging from concatenative approaches [3] to hidden Markov models [51], switching linear dynamic systems [43], restricted Boltzmann machines [52], Gaussian processes [60], and random forests [32].

Recently, Recurrent Neural Networks (RNN) have been applied to short [14, 37] and long-term prediction [66]. Convolutional networks [22] and feed-forward networks [23] have been successfully applied to long-term generation of locomotion. Early work took great care in choosing a model expressing the inter-dependence between joints [26], while recent work favors universal approximators [22, 23, 37]. Beside choosing the neural architecture, framing the pose prediction task is equally important. In particular, defining input and output variables, their representation as well as the loss function used for training are particularly impactful, as we show in our experiments.

As for quaternions in neural networks, [15] proposes a hyper-complex extension of complex-valued convolutional neural networks, and [30] presents a variation of resilient backpropagation in quaternionic domain.

### 2.1   Joint Rotations versus Positions

Human motion is represented as a sequence of human poses. Each pose can be described through body joint positions, or through 3D-joint rotations which are then integrated via forward kinematics. For motion prediction, one can consider predicting either rotations or positions with alternative benefits and trade-offs.

The prediction of rotations allows using a parameterized skeleton [14, 43, 52]. Skeleton constraints avoid prediction errors such as non-constant bone lengths or motions outside an articulation range. However, rotation prediction is often paired with a loss that averages errors over joints which gives each joint the same weight. This ignores that the prediction errors of different joints have varying impact on the body, e.g. joints between the trunk and the limbs typically impact the pose more than joints at the end of limbs, with the root joint being the extreme case. This type of loss can therefore yield a model with spurious large errors on important joints, which severely impact generation from a qualitative perspective.

The prediction of joint positions minimizes the averaged position errors over 3D points, and as such does not suffer from this problem. However, this strategy does not benefit from the parameterized skeleton constraints and needs its prediction to be reprojected onto a valid configuration to avoid issues like bone stretching [22, 23]. This step can be resource intensive and is less efficient in terms of model fitting. When minimizing the loss, model fitting ignores that the prediction will be reprojected onto the skeleton, which often increases the loss. Also, the projection step can yield discontinuities in time if not performed carefully.

For both positions and rotations, one can consider predicting velocities (i.e. deltas w.r.t. time) instead of absolute values [37, 53]. The density of velocities is concentrated in a smaller range of values, which helps statistical learning. However, in practice velocities tend to be unstable in long-term tasks, and generalize worse due to accumulation errors. Noise in the training data is also problematic with velocities: invalid poses introduce large variations which can yield unstable models.

Alternatively to the direct modeling of joint rotations/positions, physics-inspired models of the human body have also been explored [33] but such models have been less popular for generation with the availability of larger motion capture datasets [1, 39].

## 2.2 Learning a Stochastic Process

Human motion is a stochastic process with a high level of uncertainty. For a given past, there will be multiple likely sequences of future frames and uncertainty grows with duration. This makes training for long-term generation challenging since recorded frames far in the future will capture only a small fraction of the probability mass, even according to a perfect model.

Like other stochastic processes [7, 54, 55], motion modeling is often addressed by training transition operators, also called auto-regressive models. At each time step, such a model predicts the next pose given the previous poses. Typically, training such a model involves supplying recorded frames to predict the next recorded target. This strategy – called teacher forcing – does not expose the model to its own errors and prevents it from recovering from them, a problem known as *exposure bias* [46, 61]. To mitigate this problem, previous work suggested to add noise to the network inputs during training [14, 16]. Alternatively, [37] forgoes teacher forcing and always inputs model predictions. This strategy however can yield slow training since the loss can be very high on long sequences.

Due to the difficulty of long-term prediction, previous work has considered decomposing this task hierarchically. For locomotion, [22] proposes to subdivide the task into three steps: define the character trajectory, annotate the trajectory with footsteps, generate pose sequences. The neural network for the last step takes trajectory and speed data as input. This strategy makes the task simpler since the network is relieved from modeling the uncertainty due to the trajectory and walk cycle drift. [23] considers a network which computes different sets of weights according to the phase in the walk cycle.

## 2.3 Pose & Video Forecasting

Forecasting is an active topic of research beyond the prediction of human pose sequences. Pixel-level prediction using human pose as an intermediate variable has been explored [56, 59]. Related work also include the forecasting of locomotion trajectories [28], human instance segmentation [36], or future actions [31]. Other types of conditioning have also been explored for predicting poses: for instance, [47] explores generating skeleton pose sequences of music players from audio, [9] aims at predicting future pose sequences from static images. Also relevant is the prediction of 3D

(a) Architecture for short-term prediction

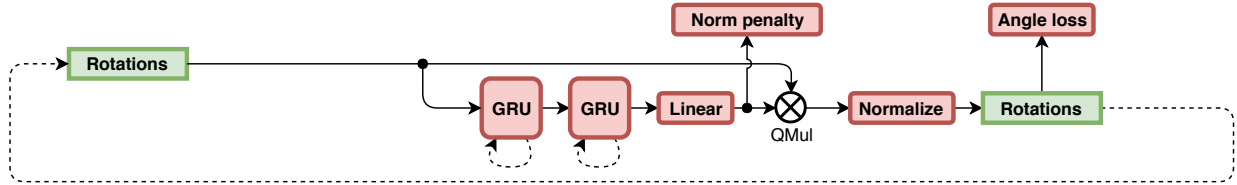

(b) Architecture for long-term generation

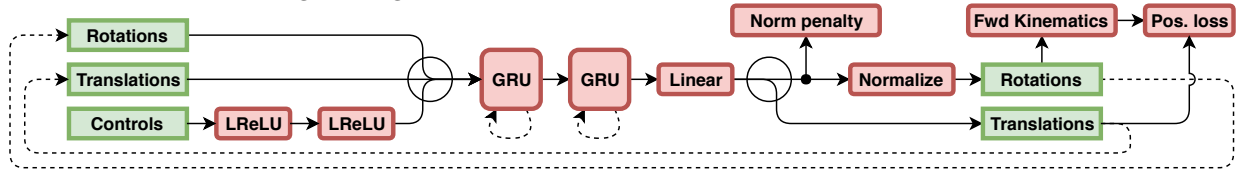

Figure 1: Architectures. "QMul" stands for quaternion multiplication: if included, it forces the model to output velocities; if bypassed, the model emits absolute rotations.

poses from images or 2D joint positions [2, 42, 45], and motion forecasting of rigid bodies. For instance, [8] models object dynamics using a neural network that performs spatial transformations on point clouds.

In terms of neural generative modeling, adversarial networks [35, 38, 56, 59] and variational auto-encoders [58, 59] are explored to explicitly deal with the intrinsic uncertainty of forecasting tasks.

## 3   QuaterNet

### 3.1   Model Architecture & Training Algorithm

We model sequences of three-dimensional poses with an RNN, as in [14, 37]. Our strategy relies on two-layer *gated recurrent unit* (GRU) networks [10]. It is an autoregressive model, i.e. at each time step, the model takes as input the previous recurrent state as well as features describing the previous pose to predict the next pose. Similar to [37], we selected GRU for their simplicity and efficiency. In line with the findings of [11], we found no benefit in using *long short-term memory* (LSTM), which require learning extra gates. Contrary to [37], however, we found an empirical advantage of adding a second recurrent layer, but not a third one. The two GRU layers comprise $1,000$ hidden units each, and their initial states $\mathbf{h_0}$ are learned from the data.

Figure 1 shows the high-level architecture of our *pose network*, which we use for both short-term prediction and long-term generation. If employed for the latter purpose, the model includes additional inputs (referred to as "Translations" and "Controls" in the figure), which are used to provide artistic control. The network takes as input the rotations of all joints (encoded as unit quaternions, a choice that we motivate in section 3.2), plus optional inputs, and is trained to predict the future states of the skeleton across $k$ time steps, given $n$ frames of initialization; $k$ and $n$ depend on the task. For learning, we use the Adam optimizer [27], clipping the gradient norm to 0.1 and decaying the learning rate exponentially with a factor $\alpha = 0.999$ per epoch. For efficient batching, we sample fixed length episodes from the training set, sampling uniformly across valid starting points. We define an epoch to be a random sample of size equal to the number of sequences.

To address the challenging task of generating long-term motion, the network is progressively exposed to its own predictions through a curriculum schedule [6]. We found the latter to be beneficial for improving the error and model stability, as we demonstrate in Figure 3(b). At every time step, we flip a coin with probability $p$ to determine whether the model should observe the ground truth or its own prediction. Initially, $p = 1$ (i.e. teacher forcing), and it decays exponentially with a factor $\beta = 0.995$ per epoch. When the model is exposed to its own prediction, the derivative of the loss with respect to its outputs sums two terms: a first term making the current prediction closer to the current target and a second term making the current prediction improve future predictions.

## 3.2 Rotation Parameterization and Forward Kinematics Loss

Euler angles are often used to represent joint rotations [18]. They present the advantage of specifying an angle for each degree of freedom, so they can be easily constrained to match the degrees of freedom of real human joints. However, Euler angles also suffer from non-uniqueness ($\alpha$ and $\alpha + 2\pi n$ represent the same angle), discontinuity in the representation space, and singularities (*gimbal lock*) [17]. It can be shown that all representations in $\mathbb{R}^3$ suffer from these problems, including the popular exponential maps [17]. In contrast, quaternions – which lie in $\mathbb{R}^4$ – are free of discontinuities and singularities, are more numerically stable, and are more computationally efficient than other representations [44]. Their advantages come at a cost: in order to represent valid rotations, they must be normalized to have unit length. To enforce this property, we add an explicit normalization layer to our network. We also include a penalty term in the loss function, $\lambda(w^2 + x^2 + y^2 + z^2 - 1)^2$, for all quaternions prior to normalization. The latter acts as a regularizer and leads to better training stability. The choice of $\lambda$ is not crucial; we found that any value between 0.1 and 0.001 serves the purpose (we use $\lambda = 0.01$). During training, the distribution of the quaternion norms converges nicely to a Gaussian with mean 1, i.e. the model learns to represent valid rotations. It is important to observe that if $\mathbf{q}$ represents a particular orientation, then $-\mathbf{q}$ (*antipodal representation*) represents the same orientation. As shown in Figure 2(b), we found these two representations to be mixed in our dataset, leading to discontinuities in the time series. For each orientation at time $t$, we enforce continuity by choosing the representation with the lowest Euclidean distance from the one in the previous frame $t - 1$ (Figure 2(c)). This representation still allows for two representations with inverted sign for each time series, which does not represent an issue in our case as we never compare quaternions directly in our loss functions.

Owing to the advantages presented above, this work represents joint rotations with quaternions. Previous work in motion modeling has used quaternions for pose clustering [63], for joint limit estimation [19], and for motion retargeting [57]. To the best of our knowledge, human motion prediction with a quaternion parameterization is a novel contribution of our work.

Discontinuities are not the only drawback of previous approaches (cf. Section 2). Regression of rotations fails to properly encode that a small error on a crucial joint might drastically impact the positional error. Therefore we propose to compute positional loss. Our loss function takes as input joint rotations and runs forward kinematics to compute the position of each joint. We can then compute the Euclidean distance between each predicted joint position and the reference pose. Since forward kinematics is differentiable with respect to joint rotations, this is a valid loss for training the network. This approach is inspired by [65] for hand tracking and [64] for human pose estimation in static images. Unlike Euler angles (used in [64, 65]), which employ trigonometric functions to compute transformations, quaternion transformations are based on linear operators [44] and are therefore more suited to neural network architectures. [57] also employs a form of forward kinematics with quaternions, in which quaternions are converted to rotation matrices to compose transformations. In our case, all transformations are carried out in quaternion space. Compared to other work with positional loss [22, 23], our strategy penalizes position errors properly and avoids re-projection onto skeleton constraints. Additionally, our differentiable forward kinematics implementation allows for efficient GPU batching and therefore only increases the computational cost over the rotation-based loss by 20%.

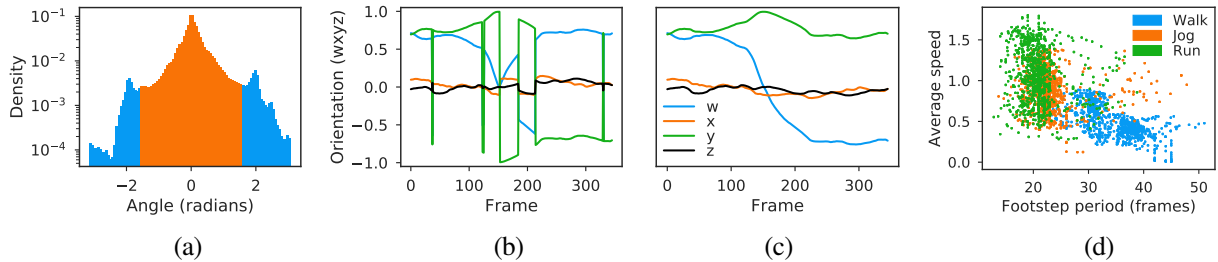

Figure 2: **(a)** Local angle distribution for H3.6m, where orange represents the safe range between $-\pi/2$ and $\pi/2$, and blue highlights the potentially problematic range (7% of all angles). **(bc)** Antipodal representation problem for quaternions. **(b)** is a real sequence from the training set, both discontinuous and ambiguous. **(c)** our approach, which corrects discontinuities but still allows for two possible choices, $q$ and $-q$. **(d)** Distribution of the gait parameters across the training set.

## 3.3 Short-Term Prediction

For short-term predictions with our quaternion network, we consider predicting either relative rotation deltas (analogous to angular velocities) or absolute rotations. We take inspiration from residual connections applied to Euler angles [37], where the model does not predict absolute angles but angle deltas and integrates them over time. For quaternions, the predicted deltas are applied to the input quaternions through quaternion product [48] (*QMul* block in Figure 1). Similar to [37], we found this approach to be beneficial for short-term prediction, but we also discovered that it leads to instability in the long-term.

Previous work evaluates prediction errors by measuring Euclidean distances between Euler angles and we precisely replicate that protocol to provide comparable results by replacing the positional loss with a loss on Euler angles. This loss first maps quaternions onto Euler angles, and then computes the L1 distance with respect to the reference angles, taking the best match modulo $2\pi$. A proper treatment of angle periodicity was not found in previous implementations, e.g. [37], leading to slightly biased results. In fact, the number of angles located around $\pm\pi$ is not negligible on the dataset we used for our experiments, see Figure 2(a).

## 3.4 Long-Term Generation

For long-term generation, we restrict ourselves to locomotion actions. We define our task as the generation of a pose sequence given an average speed and a ground trajectory to follow. Such a task is common in computer graphics [4, 13, 40].

We decompose the task into two steps: we start by defining some parameters along the trajectory (facing direction of the character, local speed, frequency of footsteps), then we predict the sequence of poses. The trajectory parameters can be manually defined by the artist, or they can be fitted automatically via a simple *pace network*, which is provided as a useful feature for generating an animation with minimal effort. The second step is addressed with our quaternion recurrent network (*pose network*).

The pace network is a simple recurrent network with one GRU layer with 30 hidden units. It represents the trajectory as a piecewise linear spline with equal-length segments [50] and performs its recursion over segments. At each time step, it receives the spline curvature and the previous hidden state. It predicts the character facing direction relative to the spline tangent (which can be used for making the character walk sideways, for instance), the frequency of its footsteps, and its local speed, which is a low-pass filtered version of the instantaneous speed on the training set. We found the two dimensions (frequency and speed) necessary to describe the character's gait (e.g. walk, jog, run), as illustrated in Figure 2(d).

This network is trained to minimize the mean absolute error (MAE) of its features. Depending on the scenario – offline or online – we propose two versions of this network: one based on a bidirectional architecture, and one based on a regular 1-directional RNN whose outputs are delayed by a small distance. The latter is particularly suitable for real-time applications, since it does not observe the trajectory far in the future.

The pose network is similar to the network we used for short-term predictions but presents additional inputs and outputs, i.e. the *Translations* and *Controls* blocks in Figure 1. The *Controls* block consists of the tangent of the current spline segment as a 2D versor, the facing direction as a 2D versor, the local longitudinal speed along the spline, and the walk cycle. The last two features are merged into a signal of the form $A[\cos(\theta), \sin(\theta)]$, where $A$ is the longitudinal speed, and $\theta$ is a cyclic signal where $0 = 2\pi$ corresponds to a left foot contact and $\pi$ corresponds to a right foot contact. For training, we extract these features from training recordings by detecting when the speed of a foot falls to zero. At inference, we integrate the frequency to recover $\theta$. Since this block is not in the recurrent path, we pass its values through two fully connected layers with 30 units each and Leaky ReLU activations (with leakage factor $a = 0.05$). We use leaky activations to prevent the units from dying, which may represent a problem with such a small layer size. The pose network also takes the additional outputs from the previous time-step (*Translations* block). These outputs are the height of the character root joint and the positional offset on the spline compared to the position obtained by integrating the average speed. The purpose of the latter is to model the high-frequency details of movement, which helps with realism and foot sliding. We extract this feature from the training set by low-pass filtering the speed along the trajectory (which yields the average local speed), subtracting the latter from the overall speed (which yields a high-pass-filtered series), and integrating it. The pose network is trained to minimize the Euclidean distance to the reference pose with the forward kinematic positional loss introduced in Section 3.2. As before, we regularize non-normalized quaternion outputs to stay on the unit-norm ball.

# 4 Experiments

We perform two types of evaluation. We evaluate short-term prediction of human motion over different types of actions using the benchmark setting evaluating angle prediction errors on Human3.6m data [14, 34, 37]. We also conduct a human study to qualitatively evaluate the long-term generation of human locomotion [22, 23] since quantitative generation of long-term prediction is difficult. For the latter, we use the same dataset as [21, 22], instead of Human3.6m.

## 4.1 Short-Term Prediction

We follow the experimental setup of [14] on the Human3.6m task [24, 25]. This dataset consists of motion capture data from seven actors performing 15 actions. The skeleton is represented with 32 joints recorded at 50 Hz, which we down-sample to 25 Hz keeping both even/odd versions of the data for training as in [37]. Our evaluation measures the Euclidean distance between predicted and measured Euler angles, similarly to [14, 34, 37]. We use the same split between train and test. We compare to previous neural approaches [14, 34, 37] and simple baselines [37], i.e. running average over 2 and 4 frames, zero-velocity (i.e. last known frame) predictions.

We train a single model for all actions, conditioning the generator on $n = 50$ frames (2 seconds) and predicting the next $k = 10$ frames (400 ms). The model is not given any action category as input. We report results both for modeling velocities (relative rotations) or absolute rotations. Table 1 shows that our method achieves the best results for all actions (walking, eating, smoking, and discussions) and time horizons compared in previous work. It highlights that velocities generally perform better than absolute rotations for short-term predictions, and that our approach performs consistently better than the state-of-the-art.

| milliseconds | Walking | | | | Eating | | | | Smoking | | | | Discussion | | | |
|---|---|---|---|---|---|---|---|---|---|---|---|---|---|---|---|---|
| | 80 | 160 | 320 | 400 | 80 | 160 | 320 | 400 | 80 | 160 | 320 | 400 | 80 | 160 | 320 | 400 |
| Run. avg. 4 [37] | 0.64 | 0.87 | 1.07 | 1.20 | 0.40 | 0.59 | 0.77 | 0.88 | 0.37 | 0.58 | 1.03 | 1.02 | 0.60 | 0.90 | 1.11 | 1.15 |
| Run. avg. 2 [37] | 0.48 | 0.74 | 1.02 | 1.17 | 0.32 | 0.52 | 0.74 | 0.87 | 0.30 | 0.52 | 0.99 | 0.97 | 0.41 | 0.74 | 0.99 | 1.09 |
| Zero-velocity [37] | 0.39 | 0.68 | 0.99 | 1.15 | 0.27 | 0.48 | 0.73 | 0.86 | _0.26_ | _0.48_ | _0.97_ | _0.95_ | _0.31_ | _0.67_ | _0.94_ | _1.04_ |
| ERD [14] | 0.93 | 1.18 | 1.59 | 1.78 | 1.27 | 1.45 | 1.66 | 1.80 | 1.66 | 1.95 | 2.35 | 2.42 | 2.27 | 2.47 | 2.68 | 2.76 |
| LSTM-3LR [14] | 0.77 | 1.00 | 1.29 | 1.47 | 0.89 | 1.09 | 1.35 | 1.46 | 1.34 | 1.65 | 2.04 | 2.16 | 1.88 | 2.12 | 2.25 | 2.23 |
| SRNN [34] | 0.81 | 0.94 | 1.16 | 1.30 | 0.97 | 1.14 | 1.35 | 1.46 | 1.45 | 1.68 | 1.94 | 2.08 | 1.22 | 1.49 | 1.83 | 1.93 |
| GRU unsup. [37] | _0.27_ | _0.47_ | _0.70_ | _0.78_ | 0.25 | 0.43 | 0.71 | 0.87 | 0.33 | 0.61 | 1.04 | 1.19 | _0.31_ | 0.69 | 1.03 | 1.12 |
| GRU sup. [37] | 0.28 | 0.49 | 0.72 | 0.81 | _0.23_ | _0.39_ | _0.62_ | _0.76_ | 0.33 | 0.61 | 1.05 | 1.15 | _0.31_ | 0.68 | 1.01 | 1.09 |
| *QuaterNet* absolute | 0.26 | 0.42 | 0.67 | 0.70 | 0.23 | 0.38 | 0.61 | 0.73 | 0.32 | 0.52 | **0.92** | **0.90** | 0.36 | 0.71 | 0.96 | 1.03 |
| *QuaterNet* velocity | **0.21** | **0.34** | **0.56** | **0.62** | **0.20** | **0.35** | **0.58** | **0.70** | **0.25** | **0.47** | 0.93 | **0.90** | **0.26** | **0.60** | **0.85** | **0.93** |

Table 1: Mean angle error for short-term motion prediction on Human 3.6M for different actions: simple baselines (top), previous RNN results (middle), our contribution (bottom). Bold indicates the best result, underlined indicates the previous state-of-the-art.

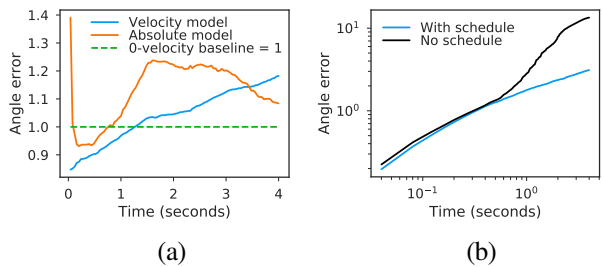

(a)                                    (b)

Figure 3: Comparison between models for a longer time span. We compare the mean angle errors for all 15 actions, each averaged over 64 test sequences. **(a)** Velocity model vs orientation model, with respect to the zero-velocity baseline (for clarity). Both models are trained with scheduled sampling. **(b)** Beneficial effect of training with scheduled sampling on the velocity model.

We also report results with a longer-term horizon on all 15 actions. Figure 3(a) shows that integrating velocities is prone to error accumulation and absolute rotations are therefore advantageous for longer-term predictions. The graph also highlights that motion becomes mostly stochastic after the 1-second mark, and that the absolute rotation model presents small discontinuities when the first frame is predicted, which corroborates the findings of [37]. Figure 3(b) reveals that if the velocity model is trained with scheduled sampling, it tends to learn a more stable behavior for long-term predictions. By contrast, the velocity model trained with regular feedback is prone to catastrophic drifts over time.

## 4.2 Long-Term Generation

Our long-term evaluation relies on the generation of locomotion sequences from a given trajectory. We follow the setting of [22]. The training set comprises motion capture data from multiple sources [1, 39, 41, 62] at 120 Hz, and is re-targeted to a common skeleton. In our case, we trained at a frame rate of 30Hz, keeping all 4 down-sampled versions of the data, and mirroring the skeleton to double the amount of data. We also applied random rotations to the whole trajectory to better cover the space of the root joint orientations. This dataset relies on the CMU skeleton [1] with 31 joints. We removed joints with constant angle, yielding a dataset with 26 joints.

Our first experiment compares loss functions. We condition the generator on $n = 60$ frames and predict the next $k = 30$ frames. Figure 4 shows that optimizing the angle loss can lead to larger position errors since it fails to properly assign credit to correct predictions on crucial joints. The angle loss is also prone to exploding gradients. This suggests that optimizing the position loss may reduce the complexity of the problem, which seems counterintuitive considering the overhead of computing forward kinematics. One possible explanation is that some postures may

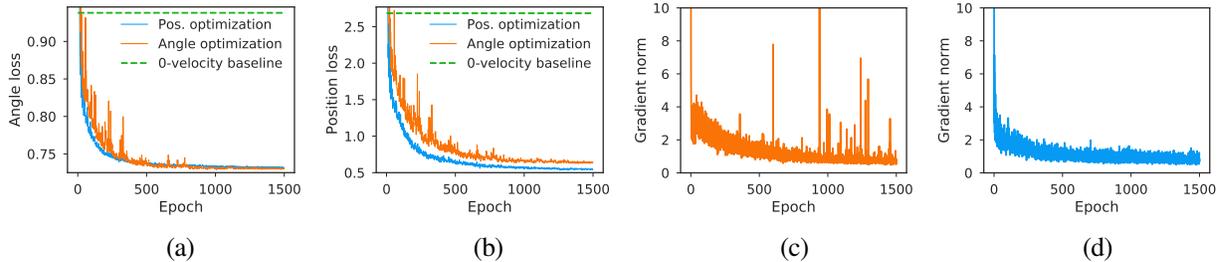

Figure 4: Training with angle versus positional loss on long-term generation. **(a)** Angle distance between joint orientations. **(b)** Euclidean distance between joint positions. Optimizing angles reduces the position loss as well, but optimizing the latter directly achieves lower errors and faster convergence. **(c)** Exploding gradients with the angle loss. **(d)** Stable gradients with the position loss. In that case, noise is solely due to SGD sampling.

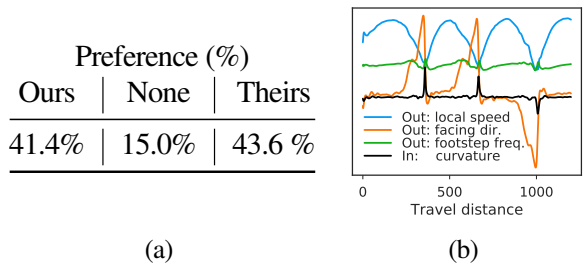

| | Preference (%) | |
|:---:|:---:|:---:|
| Ours | None | Theirs |
| 41.4% | 15.0% | 43.6 % |

(a)

(b)

Figure 5: **Left:** Human study comparing to [23]. **Right:** Our *pace network* allows fine control in space and time. Here, we instruct the character to sprint along a trajectory with sharp turns, represented as curvature spikes. The character anticipates turns by slowing down, rotating its body, and increasing the frequency of footsteps.

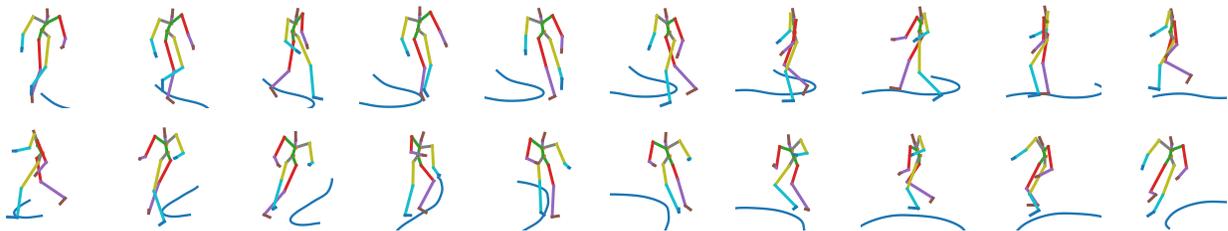

Figure 6: Example of locomotion generation. **Above:** walking. **Below:** running.

be difficult to optimize with angles, but if we consider motion as a whole, the model trained on position loss would make occasional mistakes on rotations without visibly affecting the result. Therefore, our forward kinematics positional loss is more attractive for minimizing position errors. Since this metric better reflects the quality of generation for long-term generation [22], we perform subsequent experiments with the position loss.

The second experiment assesses generation quality in a human study. We perform a side-by-side comparison with the recently proposed phase-functioned neural network [23]. For both methods, we generate 8 short clips ($\sim$ 15 seconds) for walking along the same trajectory and for each clip, we collect judgments from 20 assessors hired through Amazon Mechanical Turk. We selected only workers with "master" status. Each task compared 5 pairs of clips where methods are randomly ordered. Each task contains a control pair with an obvious flaw to exclude unreliable workers. Table 5(a) shows that our method performs similarly to [23], but without employing any post-processing.

Figure 6 shows an example of our generation where the character is instructed to walk or run along a trajectory. Figure 5(b) shows how our *pace network* computes the trajectory parameters given its curvature and a target speed. Our generation, while being online, follows exactly the given trajectory and allows for fine control of the time of passage at given way points. [22] presents the same advantages, although these constraints are imposed as an offline post-processing step, whereas [23] is online but does not support time or space constraints.

# 5   Conclusions

We propose *QuaterNet*, a recurrent neural network architecture based on quaternions for rotation parameterization – an overlooked aspect in previous work. Our experiments show the advantage of our model for both short-term prediction and long-term generation, while previous work typically addresses each task separately. Finally, we suggest training with a position loss that performs forward kinematics on a parameterized skeleton. This benefits both from a constrained skeleton (like previous work relying on angle loss) and from proper weighting across different joint prediction errors (like previous work relying on position loss). Our results improve short-term prediction over the popular Human3.6M dataset, while our long-term generation of locomotion qualitatively compares with recent work in computer graphics. Furthermore, our generation is real-time and allows better control of time and space constraints.

Our future work will apply *QuaterNet* to other motion-related tasks, such as action recognition or pose estimation from video. For motion generation, we plan to provide further artistic control with additional inputs and we would like to enable conditioning based on a rich set of actions.

Finally, to make future research in this area more easily interpretable, we suggest to report position errors instead of Euler angle errors. Alternatively, when reporting angle errors, an angle distance based on the dot product between orientations would be preferable, since it is independent of the representation. Moreover, if a particular approach regresses 3D joint positions directly, it is advisable to also report the error after reprojecting the pose to a valid skeleton.